\documentclass{article}
\usepackage{spconf,amsmath,epsfig,graphicx,subfigure}

\usepackage{algorithm}  
\usepackage{algpseudocode}  
\usepackage{amsmath}  

\begin{document}\sloppy

\def\x{{\mathbf x}}
\def\L{{\cal L}}

\title{Color Recognition for Rubik's Cube Robot}
%
\name{Shenglan Liu, Dong Jiang, Lin Feng, Feilong Wang, Zhanbo Feng, Xiang Liu, Shuai Guo, Bingjun Li}
\address{School of Computer Science and Technology, Dalian University of Technology, Dalian 116024, China}

\maketitle

\begin{abstract}
In this paper, we proposed three methods to solve color recognition of Rubik’s cube, 
which includes one offline method and two online methods. Scatter balance \& extreme
 learning machine (SB-ELM), a offline method, is proposed to illustrate the efficiency
  of training based method. We also point out the conception of color drifting which
   indicates offline methods are always ineffectiveness and can not work well in
    continuous change circumstance. By contrast, dynamic weight label propagation 
    is proposed for labeling blocks color by known center blocks color of Rubik’s cube.
     Furthermore, weak label hierarchic propagation, another online method, is also proposed for 
      unknown all color information but only utilizes weak label of center block in 
      color recognition. We finally design a Rubik’s cube robot and construct a 
      dataset to illustrate the efficiency and effectiveness of our online methods 
      and to indicate the ineffectiveness of offline method by color drifting in our dataset.
\end{abstract}
\section{Introduction}
\label{sec:intro}

Rubik’s cube puzzle has continually been as a hot topic in intelligence competition for child/adult.
 While in service robot fields, efficient solution of Rubik’s cube puzzle is a challenge 
 for computer vision. A software scheme to solve Rubik’s cube puzzle includes detection, color
   recognition and solve method of a randomly scramble cube. Rubik’s cube puzzle can be
    also considered as a sequential manipulation problem for service robot \cite{serviceRobotICCVG06}. 
    For example, optical time-of-flight pre-touch sensor are used for grasp Rubik’s 
    cube to achieve a high precise sequential manipulation.
    
In recent years, excellent research works devote in proposing quick algorithms for solving
 Rubik’s cube puzzle. Kociemba \cite{kociemba1992close} proposed close to God’s algorithm to solve the problem. 
 Korf \cite{korf1985depth} \cite{korf2008linear} involved tree search and graph search to enhance the efficient performance of 
 Rubik’s cube puzzle. Rokicki \cite{twenty2014} et al. proposed "Rubik’s Cube Group Is Twenty", which 
 is a ground-breaking results in Rubik’s cube puzzle. The above algorithms focus on 
 solution of cube for human or robots. However, many procedures need to be completed 
 before the solution which is the last procedure in service robot for Rubik’s cube 
 puzzle. In cube service robot, we first detect the location of the cube by camera. 
 Then, color recognition method need to achieve after get the color block of cube, 
 which is one of the most important problem in the whole cube service robot procedures. 
 The whole sequential manipulation of cooperating service robots should get correct 
 color information of each surface in Rubik’s cube. In fact, we will meet an important 
 issue that incorrect color information make a wrong initialization for algorithm of 
 Rubik’s cube puzzle. However, a few works focus on color recognition of Rubik’s cube 
 puzzle to our best knowledge. Cezary \cite{MRROC2018} et al. introduce a full robot project of Rubik’s 
 cube including grasp the cube, color recognition, solving algorithm and mechanical arm 
 controlling, which uses HSV color space to realize the color block distribution and 
 illustrates the importance of color recognition in Rubik’s cube solving robot. 
 In computer vision, many image processing methods can be referred to color recognition of Rubik’s cube (CRRC) 
 problem \cite{pud2016} \cite{colorhistogram} \cite{CDH}. Although we can get numerous images of cubes, 
 it is hard to generalize all the situations caused by complex environment, e.g. illumination changing, 
 cube material and abrasion by long time using. Therefore, most pre-trained models can not achieve 
 a satisfactory performance on CRRC problem, which can be denoted as \emph{color drifting}. 
\begin{figure}
    \centering
    \includegraphics[width=7.5cm]{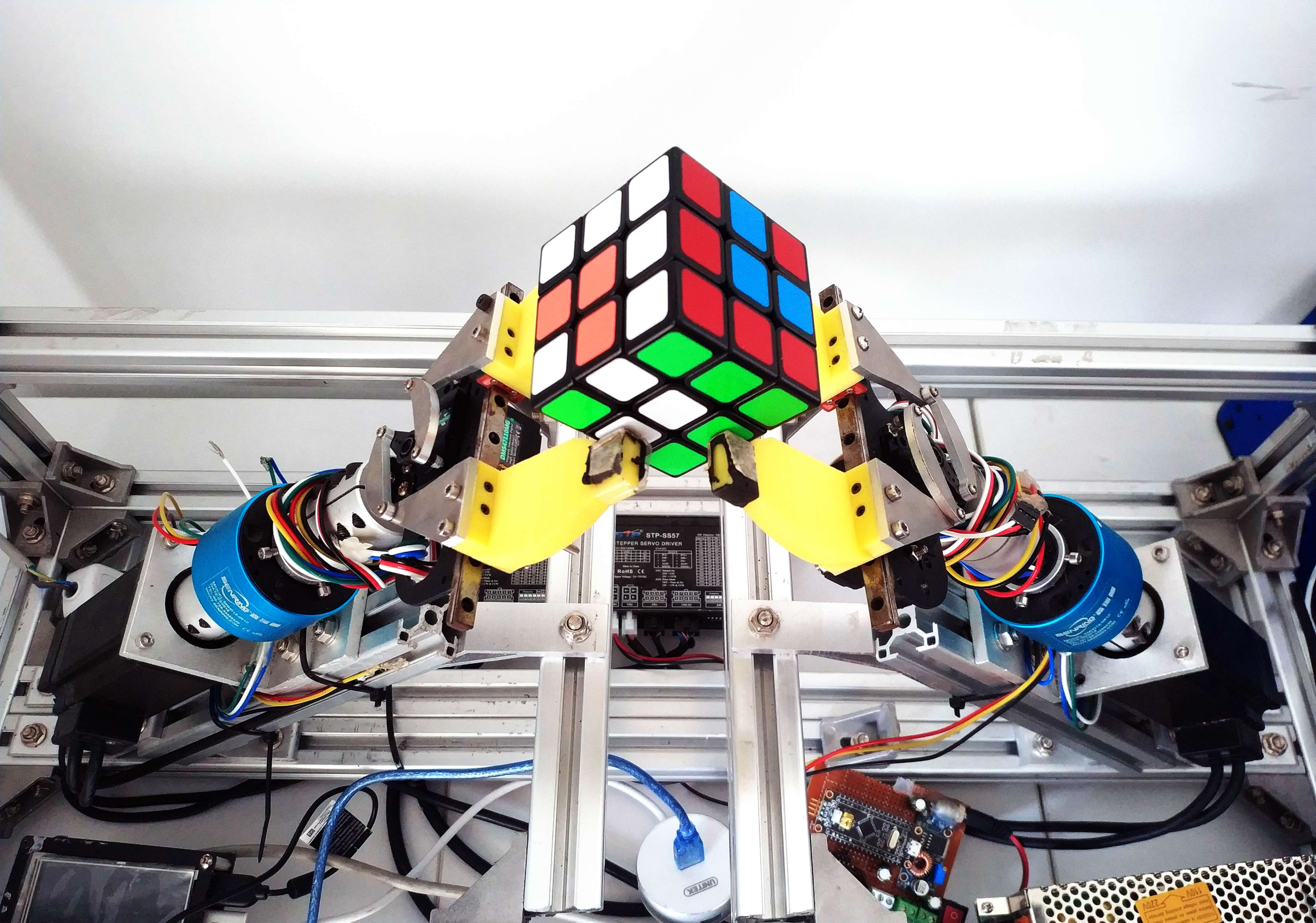} 
\caption{Cube-solving Robot}
\label{robot}
\end{figure}

In this paper, we collect different illumination and material images and label the 
images using perceptual visual interactive learning (PVIL) method \cite{liu2018perceptual} of Rubik’s cubes 
as a dataset (RC dataset) which contains 348 images and offers two HSV’s features. 
To solve CRRC problem, we introduced one model and proposed two models for CRRC 
problem including offline and online approaches. The offline method using 
supervised dimensionality reduction and classification methods calls scatter 
balance \& extreme learning machine (SB-ELM) to enhance the robustness and considers 
different utilizing environment for CRRC problem. The online methods fall into two 
approaches as follows. 

(1) Weak label hierarchic propagation (WLHP) base on weak supervised hypothesis \cite{Rajchl2016Learning}
is proposed in this paper for a special problem which indicates nine color blocks 
on each face of Rubik’s cube. 

(2) Dynamic weight label propagation (DWLP), which is motivated by idea of KNN, designs known center block 
color of each surface on Rubik’s cube. Here, we use orange and blue center block 
surfaces face to the camera as original status on robot arm (See Fig. \ref{robot}). 

The rest of this paper is organized as follows. The offline and online methods of color
 recognition on Rubik’s cube is proposed in section 2. Our new Rubik’s cube dataset 
 for color recognition is introduced and analyzed in section 3. In section 4, 
 the experimental results are listed and analyzed. The conclusion is the last section.

\section{Proposed Method}

\subsection{Overview}

In this section, we propose two HSV’s features to represent color blocks of Rubik’s cube. Based 
on HSV's features, one offline approach and two online approach are introduced as follows.
 The offline method using supervised dimensionality reduction and 
classification methods calls scatter balance \& extreme learning machine (SB-ELM). 
The online methods are Weak label hierarchic propagation (WLHP) and 
Dynamic weight label propagation (DWLP). WLHP takes the fact that none of the six central color blocks 
of Rubik's cube belong to the same color. While the process of DWLP adopts specific colors of six central blocks.
\subsection{Features}
Before color recognition, it is necessary to select suitable features to represent
 color blocks. RGB color space is often used in computers to represent color. 
 While compared to the RGB color space, HSV color space makes it easier to recognize colors because the HSV color space simulates 
 human perception of color \cite{Datta2005Content} \cite{huang1997image} \cite{zheng2014packing}. We propose two features built in the HSV color 
 space in this paper, one is 3DHSV, the other one is 16DHSV.

\textbf{3DHSV}: Since the pixels in one color block are concentrated in HSV color space, we only need to
find the coordinates of the pixel with the highest frequency in histogram as the feature. As Fig. \ref{fig3dhsv} shows, for each dimension in HSV 
color space, 3DHSV calculates the histogram on the color block and takes the value of the max bin in histogram
 as one dimension.
 
\textbf{16DHSV}: 3DHSV might ignore key information in certain situations, such as 
in the phenomenon of reflection. To improve 3DHSV, we design an uneven histogram feature based on the distribution
 of color blocks \cite{MRROC2018}. According to the color distribution characteristics, uneven histogram 
 statistics are performed on the color blocks to obtain 16DHSV features.
\subsection{Scatter balance \& extreme learning machine}
We use a novel supervised dimensionality reduction method ALDE (Angle Linear Discriminant Embedding)
 \cite{scatter_balance} to extract the features, with ELM (Extreme Learning Machine) \cite{ELM} as 
 the classifier. ALDE redefines the within-class scatter matrix and between-class scatter matrix by 
 introducing CO-Angle measurement. Thus ALDE is not inclined to neither within-class scatter nor 
 between-class scatter. ALDE can be transformed to the following optimization problem: 

\begin{figure}
    \centering
    \includegraphics[height=4cm]{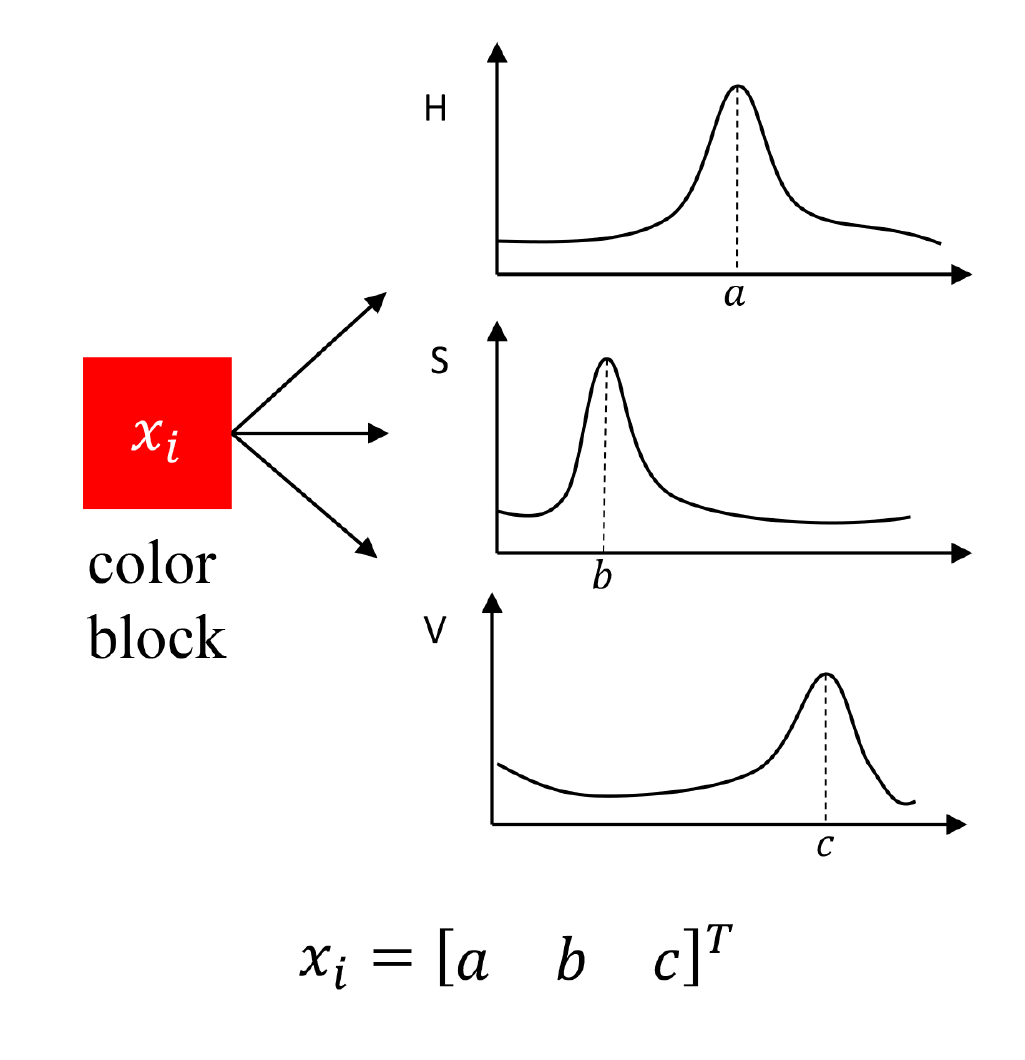}
    \caption{Description of the 3DHSV} 
    \label{fig3dhsv} 
\end{figure}

\begin{gather}
\begin{split}
\mathop {\max }\limits_W tr\left[ {W^TMW} \right] \\
M=\frac{1}{d}I-{S}'_w +{S}'_b \\
s.t. \ W^TW=I
\end{split}
\end{gather}

where ${S}'_w $ and ${S}'_b$ are the new defined within-class scatter matrix and between-class 
scatter matrix, $I$ is the identity matrix, and $W$ is the transformation matrix.
 ${S}'_w $ and ${S}'_b$ are showed as below: 

\begin{gather}
\label{SWSB}
\begin{split}
{S}'_w =\frac{1}{N}\sum\limits_{i=1}^c
{\sum\limits_{j=1}^{N_i } {(x_{ij} -\mu _i )_e (x_{ij} -\mu _i )_e^T
} } \\
{S}'_b =\frac{1}{N}\sum\limits_{i=1}^c {N_i (\mu _i -\mu )_e
(\mu _i -\mu )_e^T }
\end{split}
\end{gather}

where $c$ denotes the number of classes, $N$, $N_i$ denote the number of all samples
 and the number of samples in class $i$. In addition,  $\mu$, $\mu_i$ denote the mean of all samples
  and the mean of samples in class $i$. While $e$ denotes the process of unitizing. 

Furthermore, ELM is a feed-forward neural network with a hidden layer, 
which has extreme fast training speed and high recognition accuracy. 
ELM can be mathematically modeled as: 

\begin{gather}
\label{ELM}
\begin{split}
 \mathop{\min }\limits_\beta \frac{1}{2}\left\| \beta \right\|+\frac{1}{2}C&\sum\limits_{i=1}^N {\xi _i^2 } \\
 s.t. \ \sum\limits_{i=1}^{\tilde {N}} {\beta_i g(a_i \cdot x_j+b_j)-t_j =}&\xi _j, j=1, 2,\ldots , N 
\end{split}        
\end{gather}

where $x_j$ denote the samples, $t_i$ denote their labels, and $g(x)$ is 
the activation function. While $a_i$ denote input weights, $b_i$ denote bias parameters, 
and $\beta_i$ are output weights. A sigmoid function is used as the activation function: 

\begin{gather}
\begin{split}
g(x) = \frac{1}{1+\emph{exp}(-x)}
\end{split}
\end{gather}

\subsection{Weak label hierarchic propagation}

Weak label hierarchic propagation (WLHP) takes advantage of the sociality in 
identifying the cube state to get data with weak labels, and the labels are used 
by hierarchical propagation to solve CRRC problem.

Actually, to solve CRRC problem, it only needs to know the distribution 
information of each cube block, so CRRC problem can be simplified. 
All we have to do is just distinguishing the belonging surfaces of cube blocks, 
without knowing their specific colors. And it is a fact that 
none of the six central color blocks of the cube belong to the same color. 
So WLHP takes the features of six central color blocks as start points of propagation,
 then find other color blocks that belong to the same face with the corresponding central 
 color block by the hierarchic propagation.

 \begin{algorithm}[htbp]
    \caption{ Weak label hierarchic propagation.}  
    \label{alg:SKNN}  
    \begin{algorithmic}[1]  
      \Require  
        The set of 54 features for cube color blocks, $X$;  
      \Ensure  
        The recognition results, $L$;  
      \For{$i=0;i<6;i++$}
        \State $j=9i+5;$
        \label{algetcenter}
        \State $l_j=i$;
        \State remove $x_j$ from X;
        \label{algr1}
        \State get 2 nearest neighbors index of $x_j$ in $X$ by KNN, stored in $M_i$;
        \label{algget2nn}
        \For{$m \in M_i$}
            \State $l_m=i$;
            \State remove $x_m$ from $X$;
            \label{algr2}
        \EndFor
      \EndFor
      \For{$i=0;i<6;i++$}
      \For{$m \in M_i$}
        \State get 3 nearest neighbors index of $x_m$ in $X$ by KNN, stored in $N_i$;
        \label{algget3nn}
        \For{$n \in N_i$}
        \State $l_n=i$;
        \State remove $x_n$ from $X$;
        \label{algr3}
        \EndFor
      \EndFor
      \EndFor
    \State \Return $L$;
    \end{algorithmic}  
    \label{WLHP}
\end{algorithm}
The implementation of the hierarchic propagation is shown in Alg. \ref{WLHP}, 
where $X$ and $L$ is the set of features and labels for cube color blocks respectively. 
$x_i \in X, l_i \in L$, the subscript $i$ of $x_i$ or $l_i$ donates the index of 
cube color blocks. $M_i$ and $N_i$ record the intermediate results of 
nearest neighbor index. At the initial time, $X$ contains color block features 
of a cube obtained online, there a total of 54 features. Hierarchical propagation is to deal with the situation where the feature distribution of the central color 
block deviates from the center of the color category.

\subsection{Dynamic weight label propagation}

Label Propagation is a semi-supervised machine learning algorithm that assigns labels to 
previously unlabeled data $ $ points \cite{zhu2002learning}. Dynamic weight label propagation uses dynamic 
weights which base on labels. For CRRC problem, the size of the data is very small (54 features),
 so KNN is used as a measure of label propagation rather than constructing graph. Weight matrix is 
 used to adjust the distance between two samples, which can be described as follows:
\begin{gather}
    d_{jk}=\left\| {w_i I (x_j - x_k)} \right\|
    \label{distance}
\end{gather}

where $d_{jk}$ is the distance between $x_j$ and $x_k$, $x_j$ represents the sample with the label 
of $i$, $x_k$ represents the sample without labels, $w_i$ denotes the weight matrix of label $i$.

\begin{figure}[htbp]
    \centering
    \includegraphics[width=4cm]{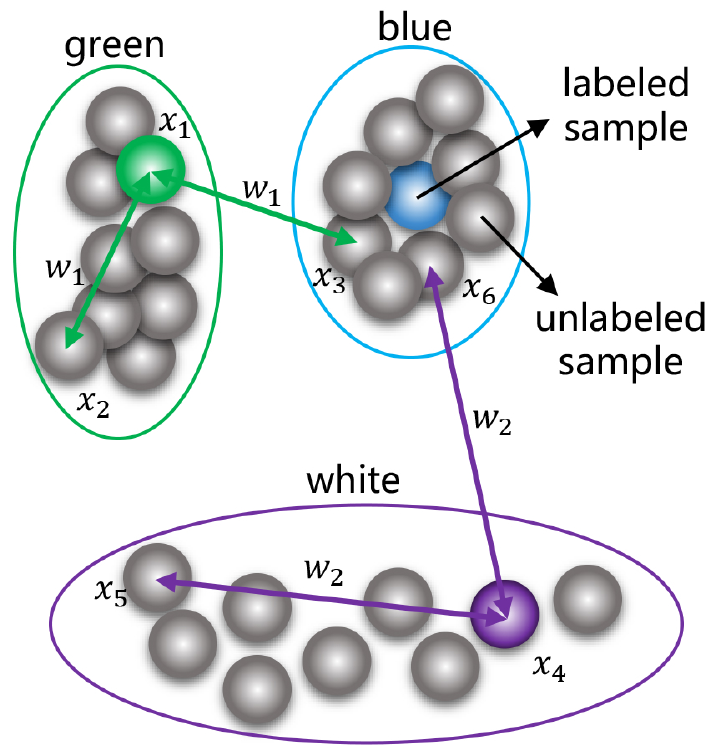} 
\caption{Dynamic weight label propagation}
\label{DWLP}
\end{figure}

As we can see in Fig. \ref{DWLP}, during the propagation process of green labels, if we directly calculate
 the distance from $x_1$ to other samples without weight matrix $w_1$, it is hard to choose $x_2$ and $x_3$.
 In DWLP, distance is calculated in the form of Eq. \ref{distance},
   it will be easy to recognize $x_2$ and   $x_3$. The matrix $w$ is dynamically determined by the label 
   during the propagation process. For example, $x_2$ is used in the propagation process of white label. 
    In this way, the distribution characteristics of the data are well utilized, and a high recognition 
    rate can be obtained.

\section{Rubik's Cube Dataset}

\subsection{Overview}
To evaluate performance of the color classification algorithm, we have collected a set of
 Rubik's cube images under different environmental circumstances. In this section, 
 we first introduce the collection process of the Rubik’s cube dataset, then explain 
 how to label the images with perceptual visual interactive learning (PVIL) method \cite{liu2018perceptual}.

\subsection{Samples collection}
\begin{figure}[htbp]
\centering
\subfigure[one group images]{
    \centering
    \includegraphics[height=5cm]{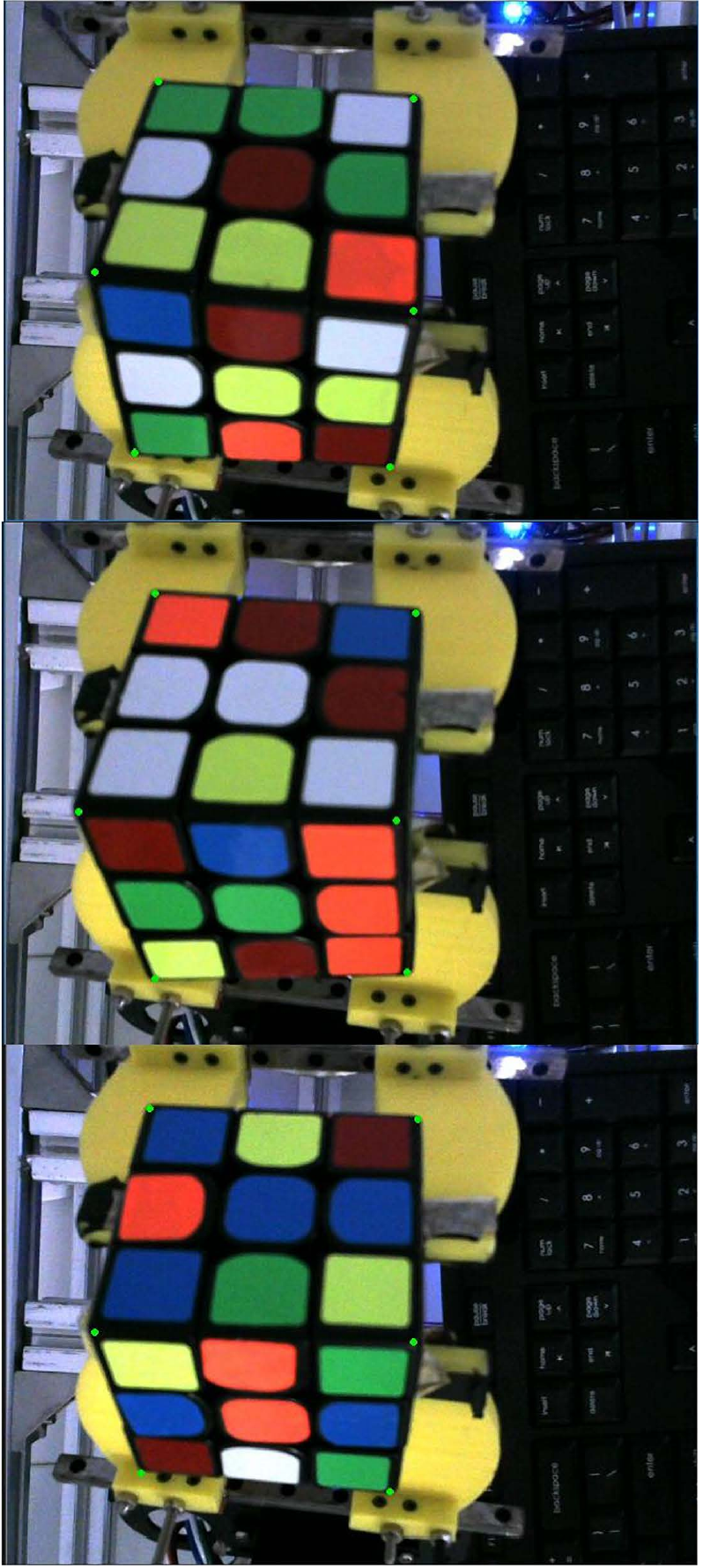} 
    \label{figtrf:a} 
} 
\hspace{0cm}
\subfigure[regular squares]{
    \centering
    \includegraphics[height=5cm]{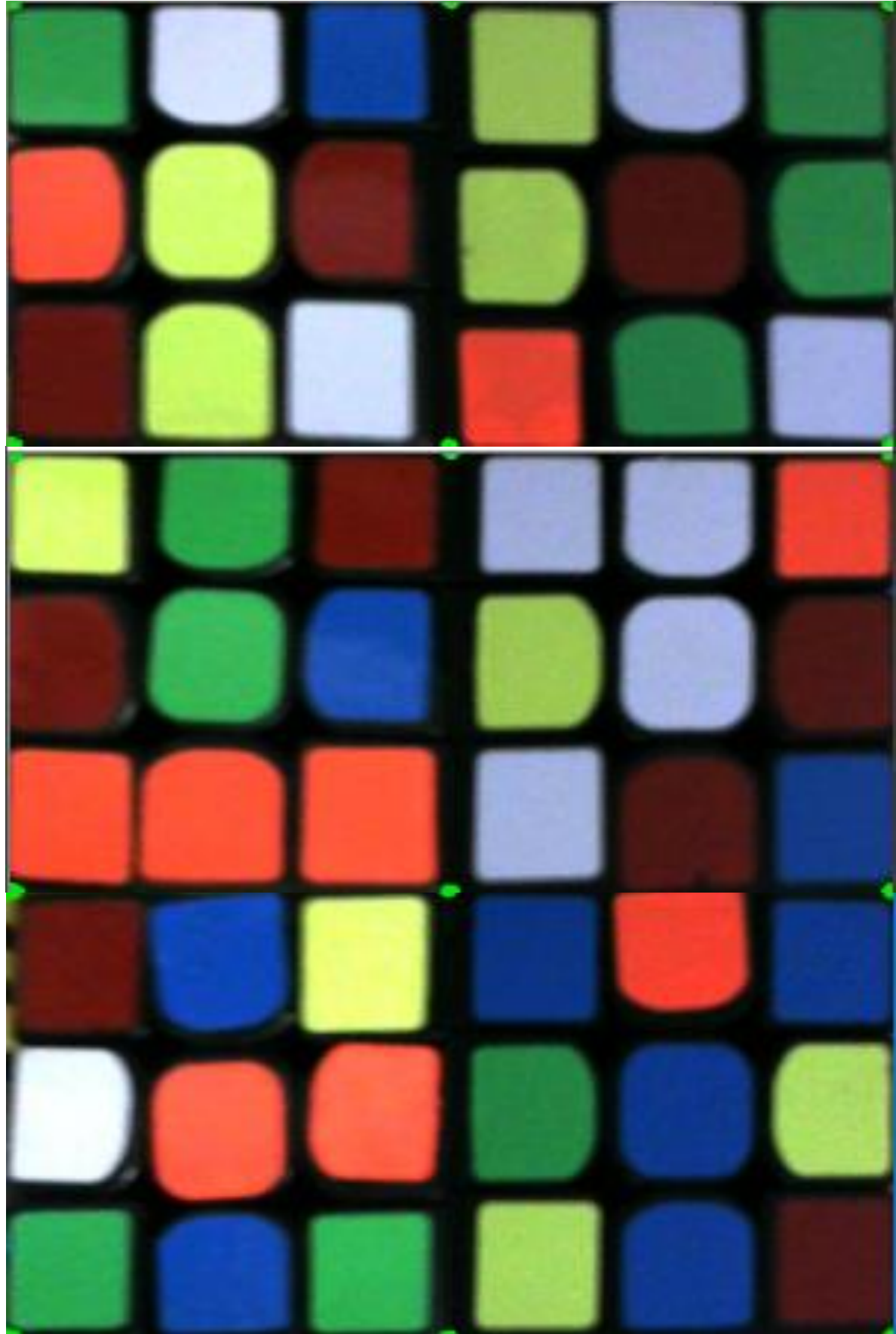}
    \label{figtrf:b}
}
\caption{Separate the cube surfaces from background}
\label{figtrf}
\end{figure}

We put the cube in the hands of the robot and use the camera to take pictures in 
different situations. Images are collected in groups. Each group represents
 the state of one cube, which consists of three images. And each image in one group
  contains two faces of the cube, as Fig. \ref{figtrf:a} shows.

In order to facilitate the extraction of color features, it is necessary to separate
  the cube form the background, so we manually mark the six corners of the cube then 
  use perspective transformation to convert irregular surfaces into regular squares, 
  as Fig. \ref{figtrf:b} shows.

\begin{figure}[htbp]
    \centering
\subfigure[cube A in bright]{
    \centering
    \hspace{0cm}
    \includegraphics[height=1.5cm]{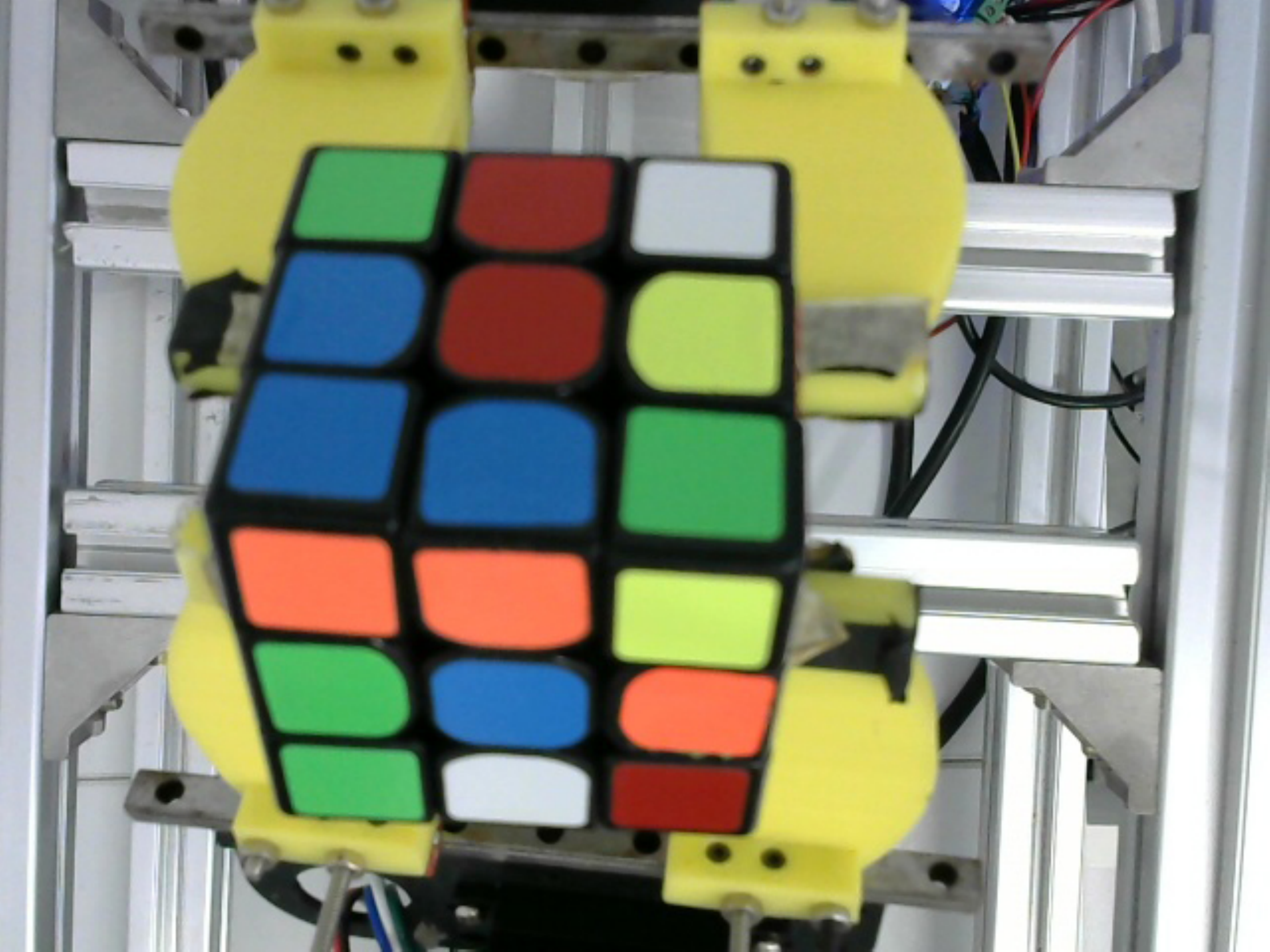}
    \includegraphics[height=1.5cm]{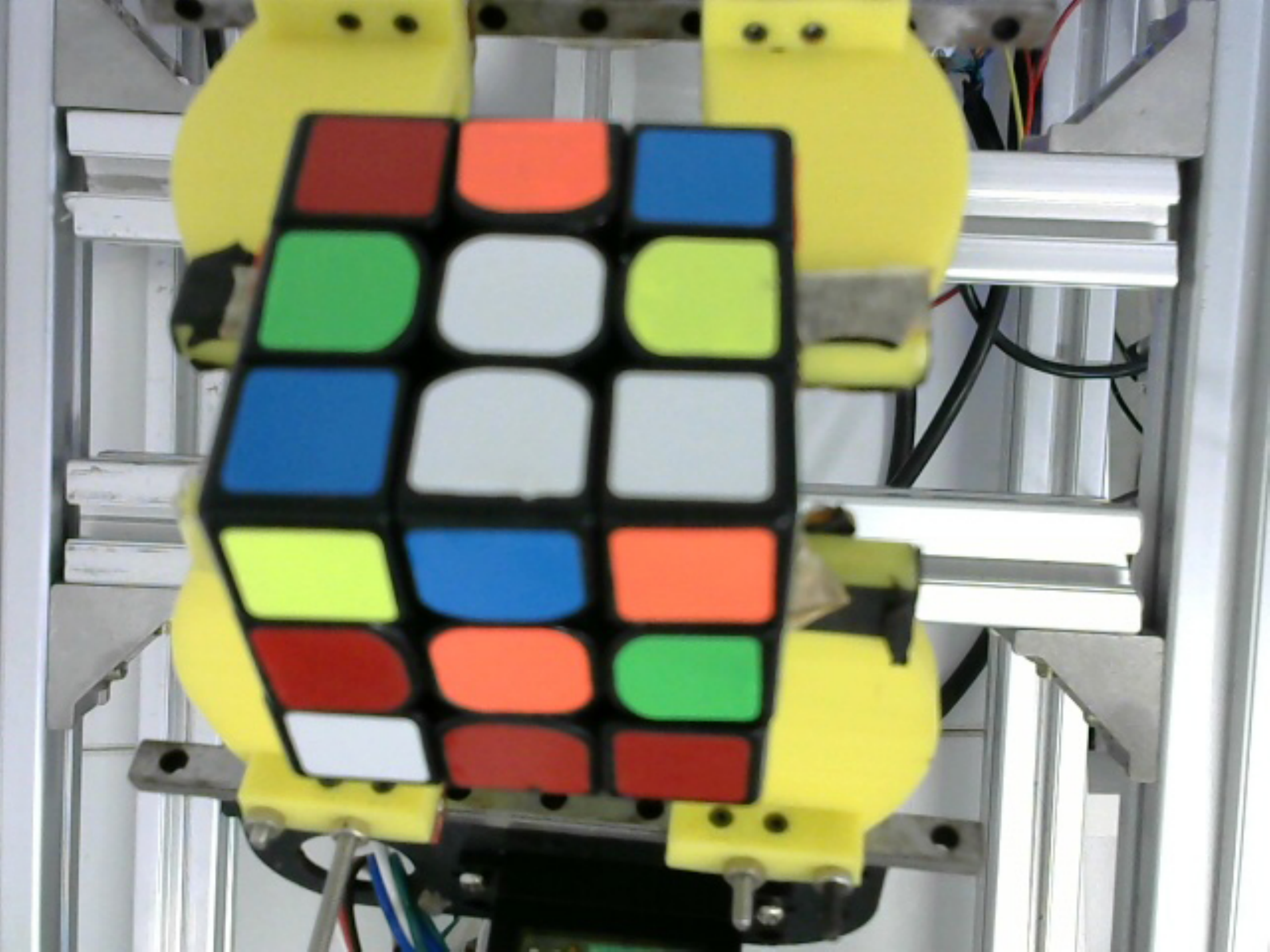}
    \includegraphics[height=1.5cm]{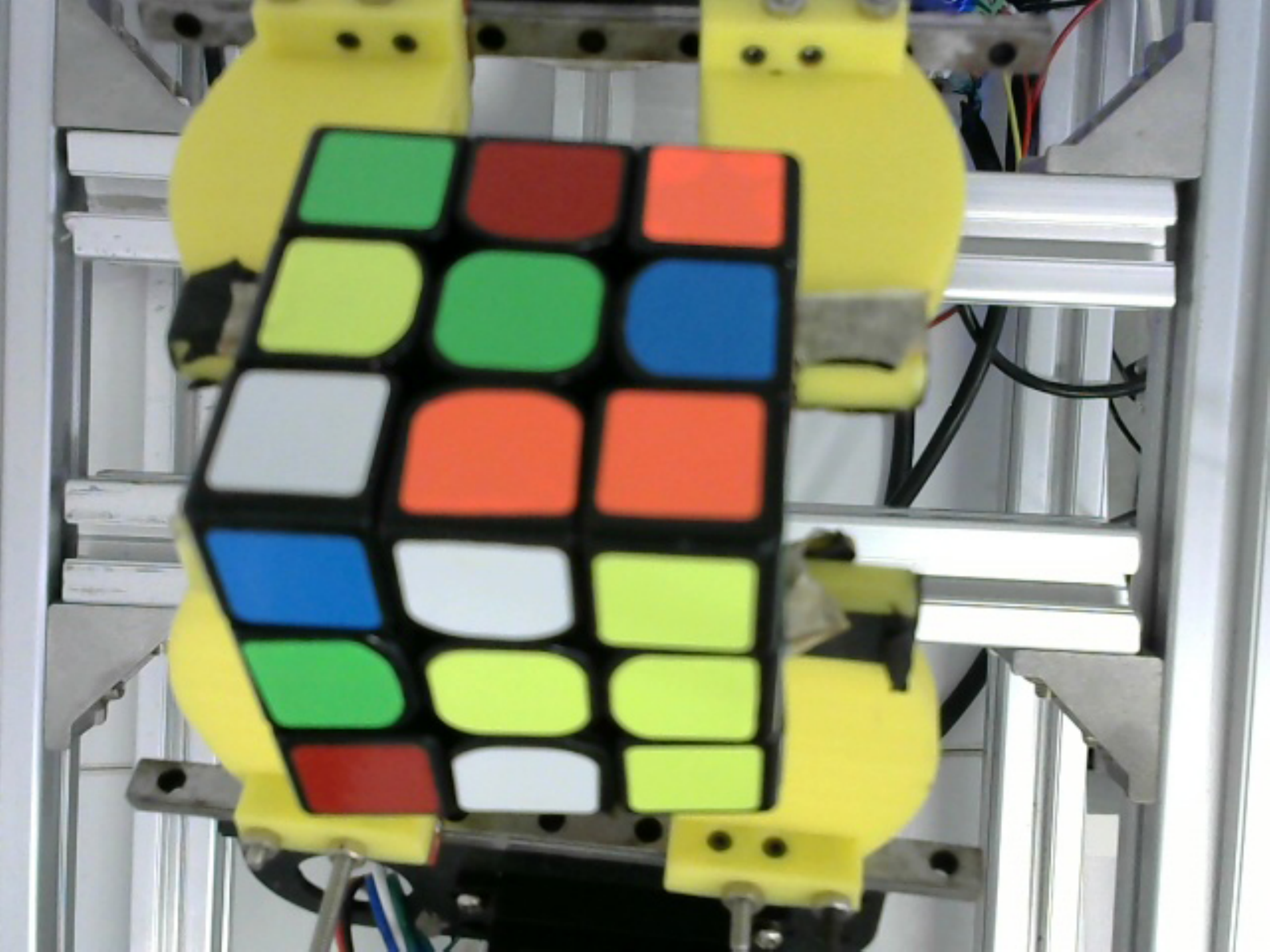}
}

\subfigure[cube A in dark]{
    \centering
    \includegraphics[height=1.5cm]{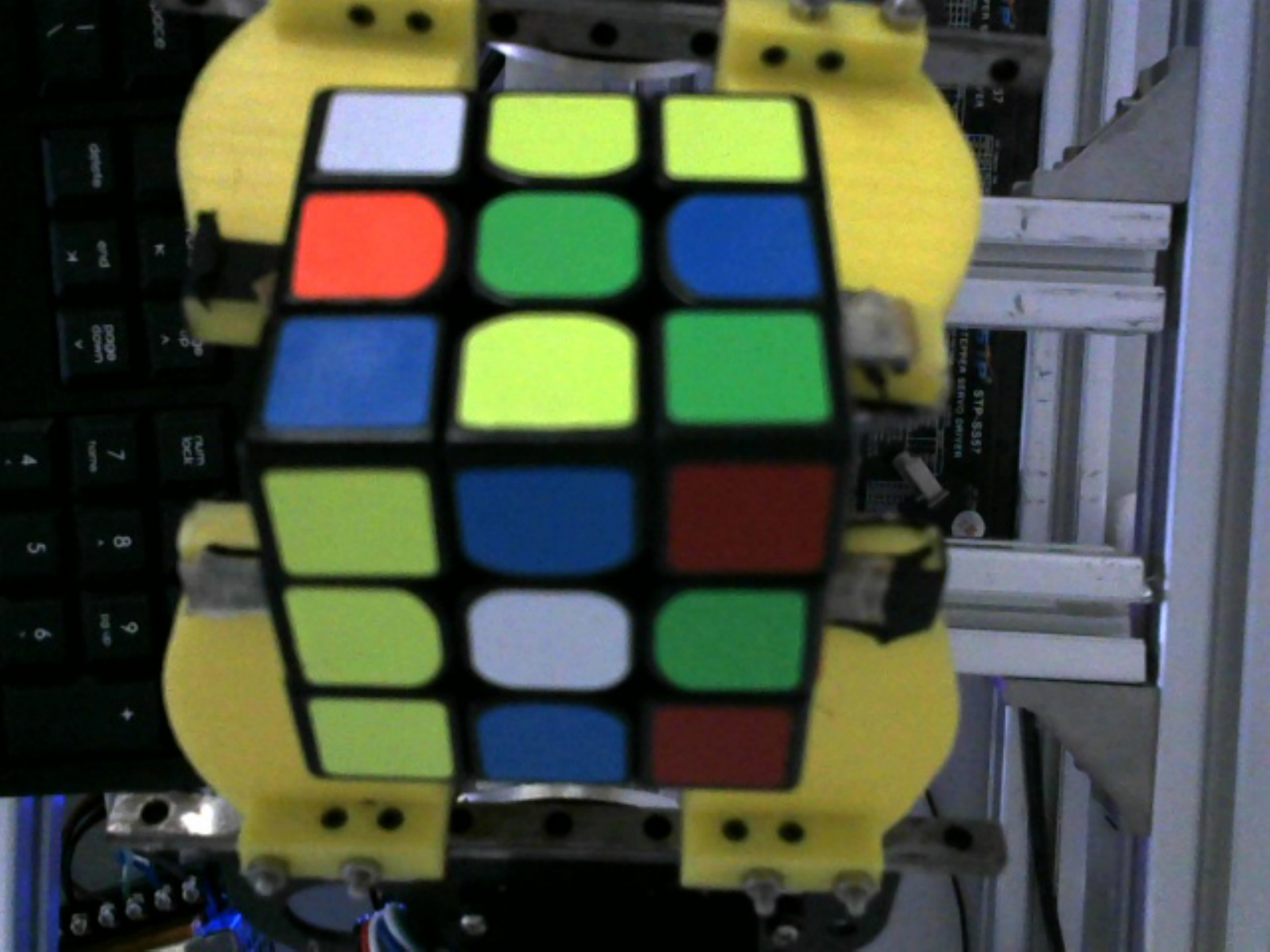}
    \includegraphics[height=1.5cm]{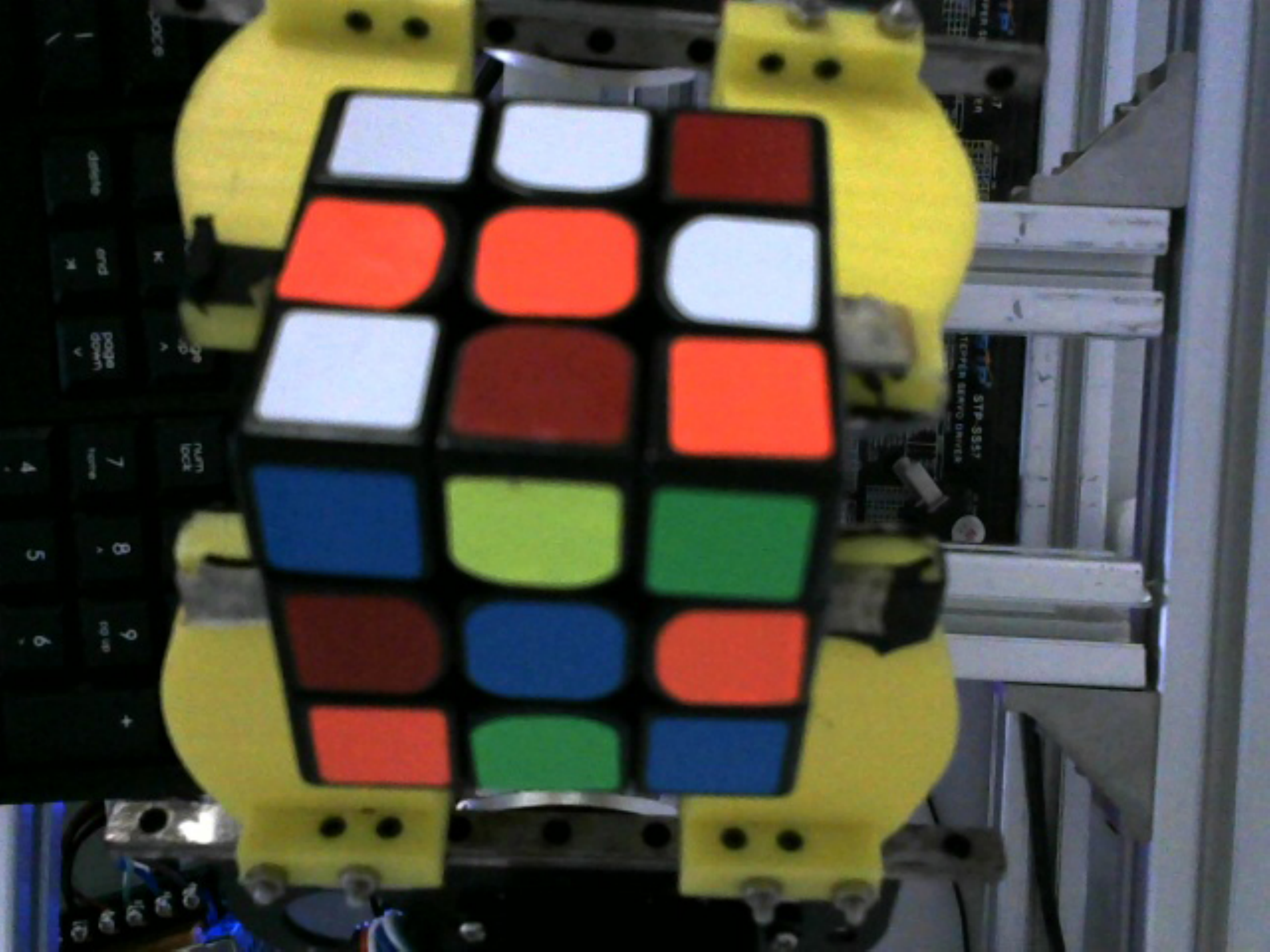}
    \includegraphics[height=1.5cm]{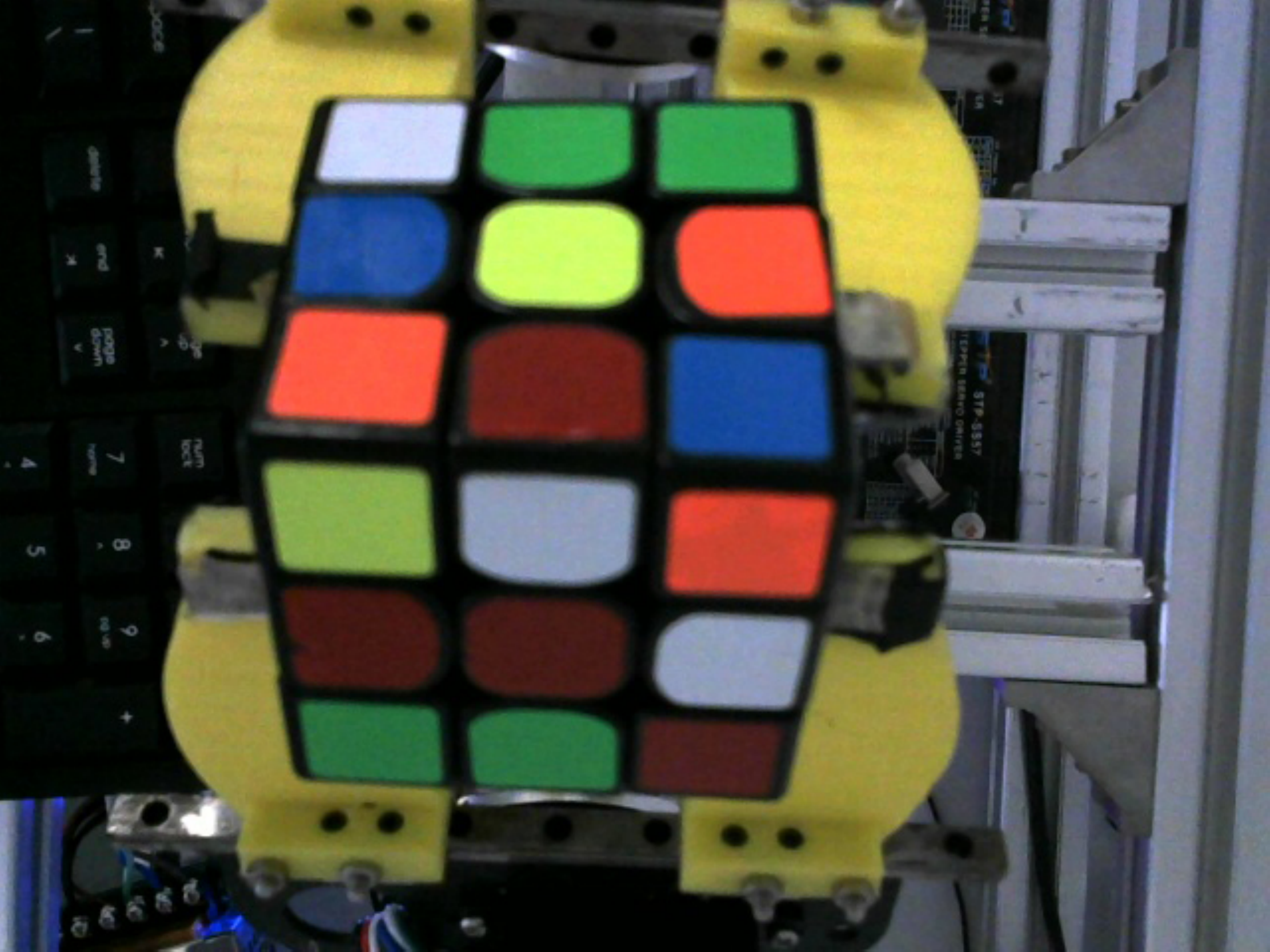}
}

\subfigure[cube B in bright]{
    \centering
    \includegraphics[height=1.5cm]{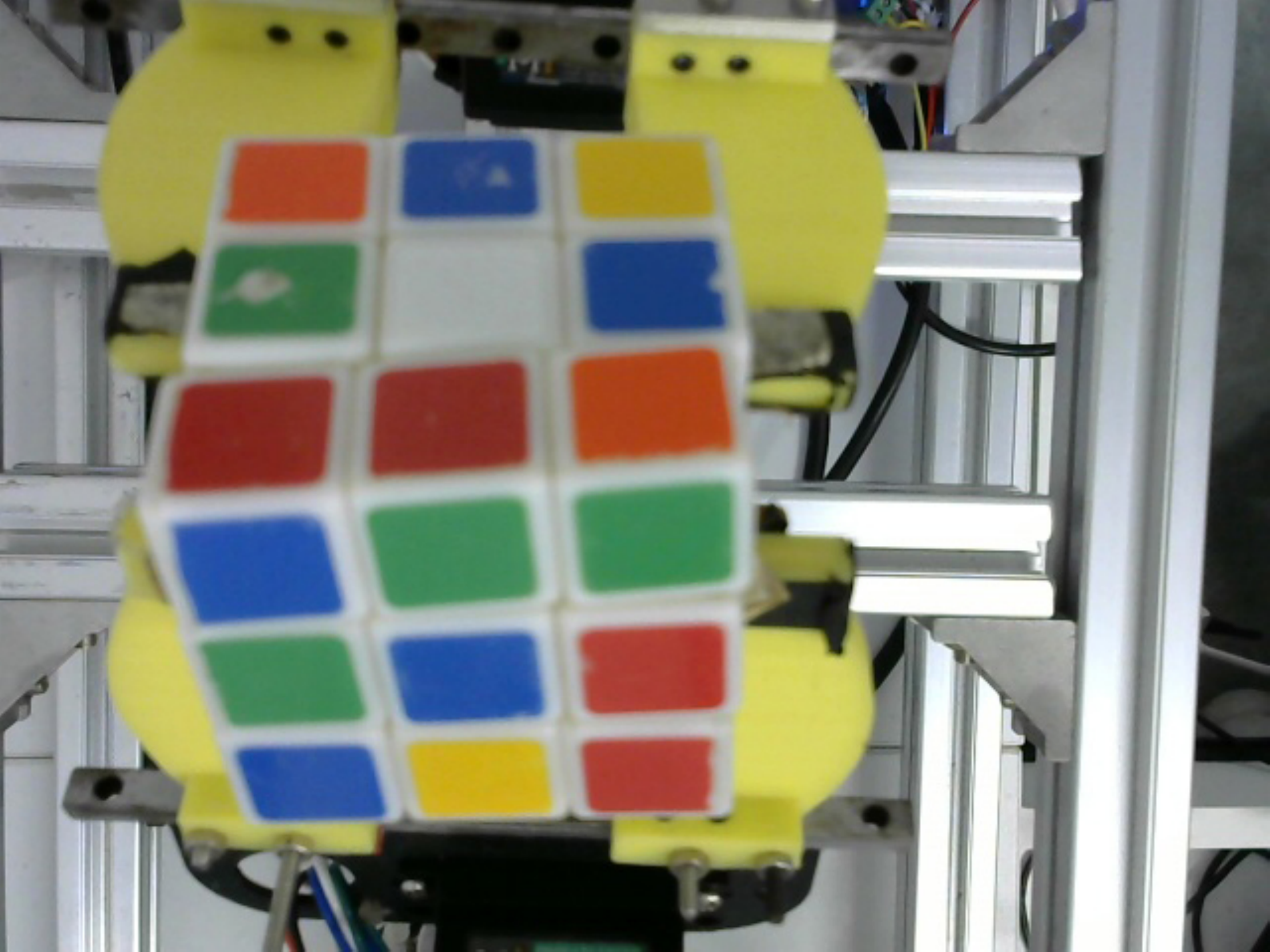}
    \includegraphics[height=1.5cm]{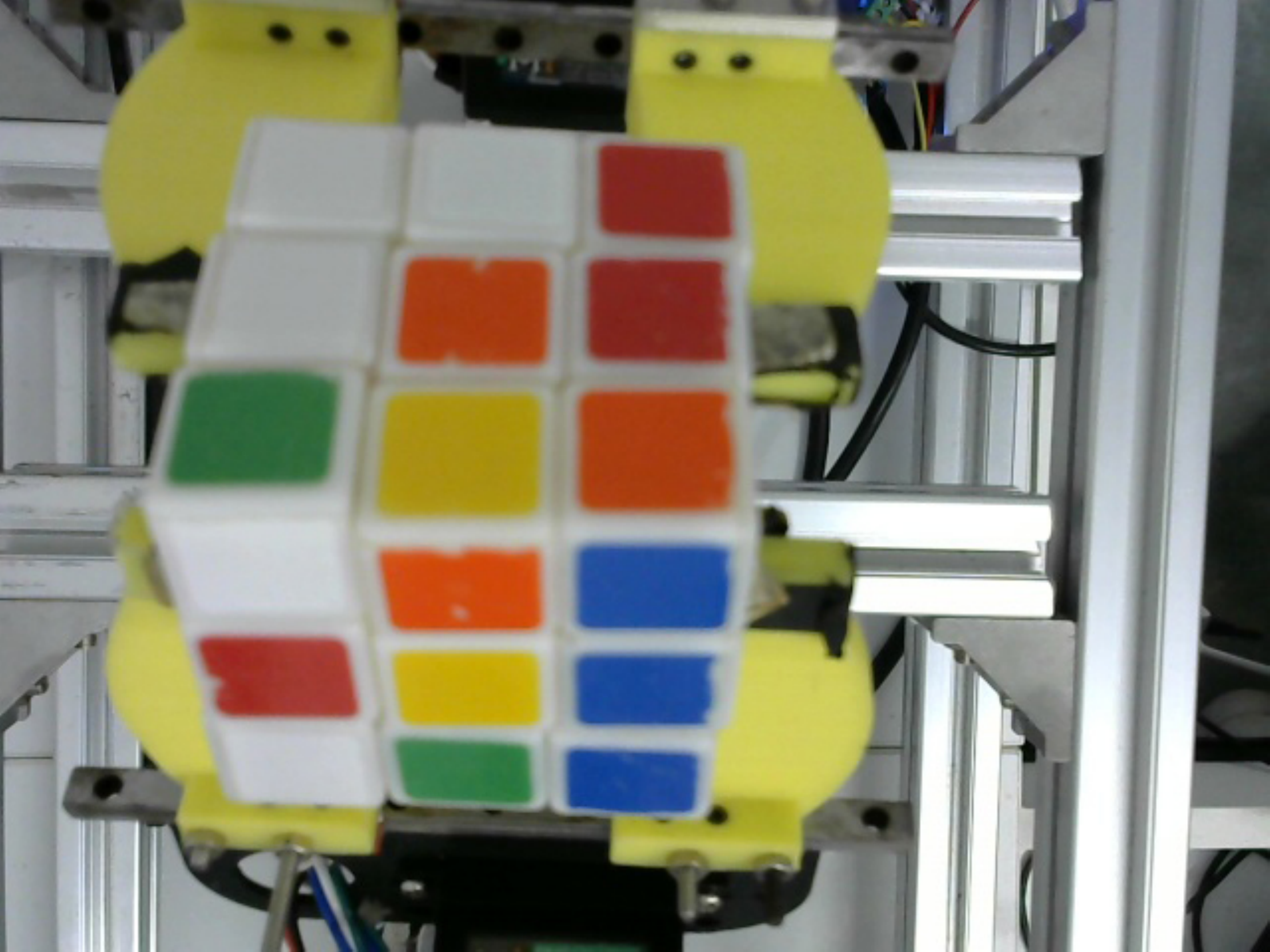}
    \includegraphics[height=1.5cm]{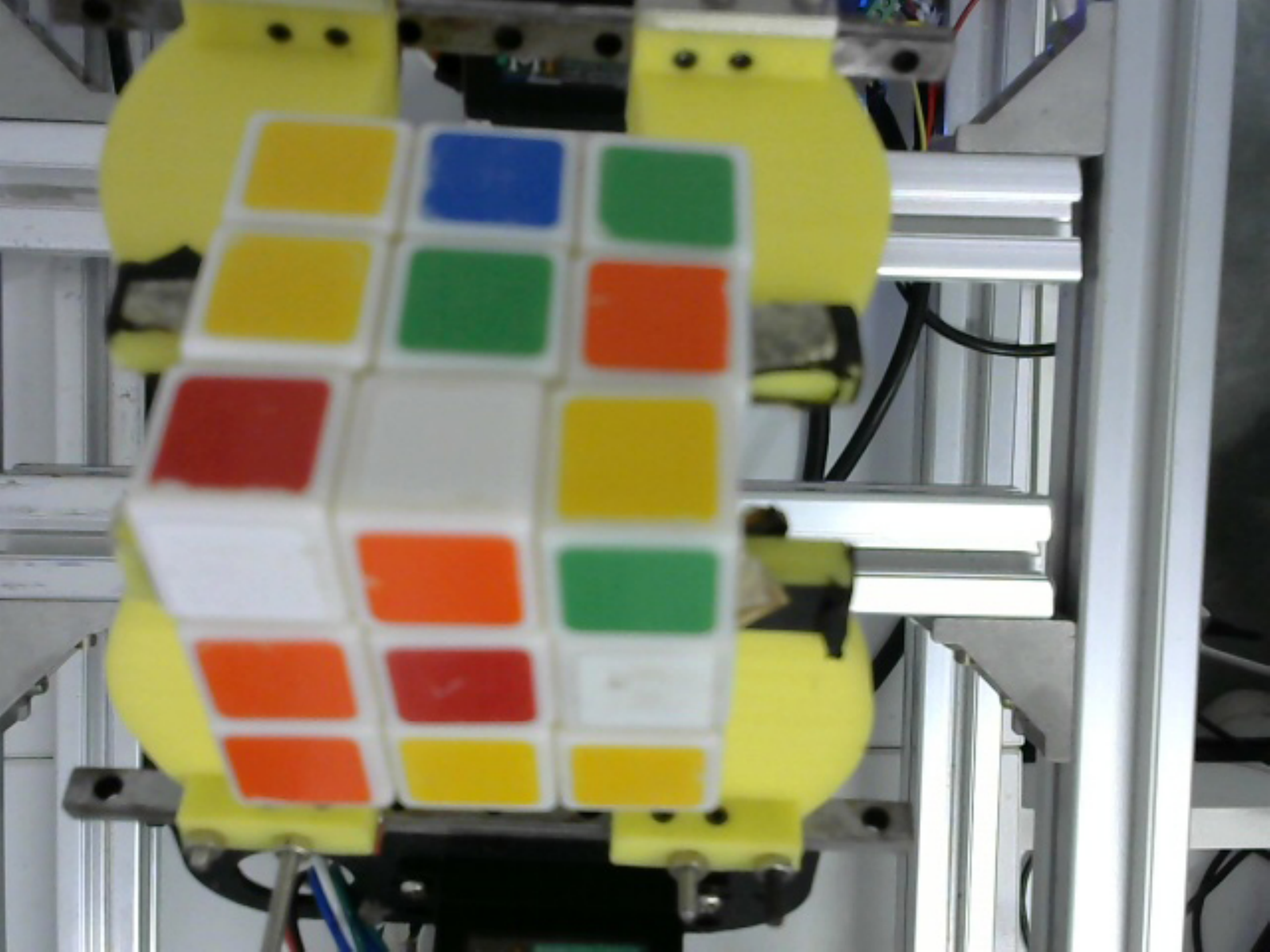}
}

\subfigure[cube B illuminated from right]{
    \centering
    \includegraphics[height=1.5cm]{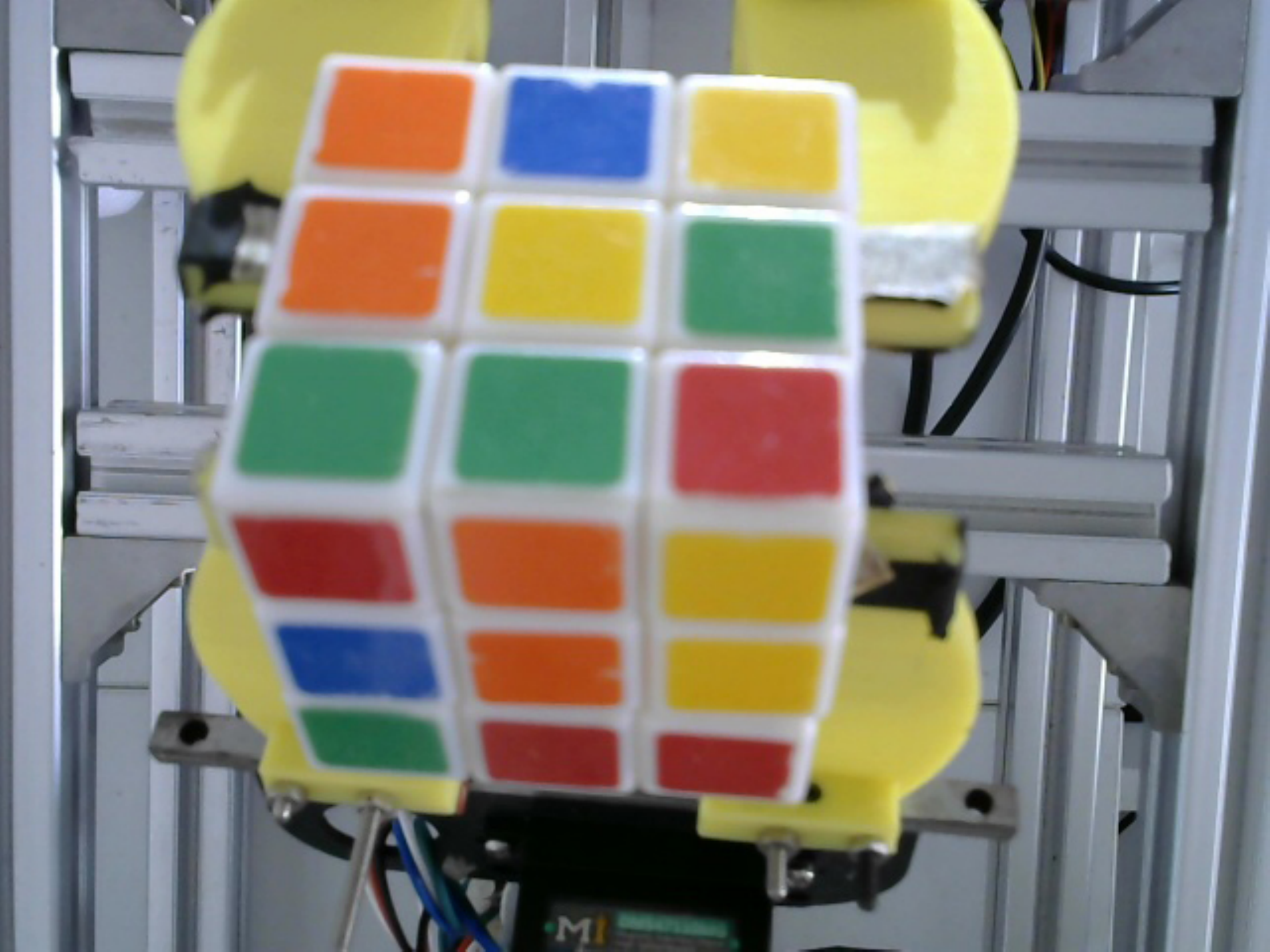}
    \includegraphics[height=1.5cm]{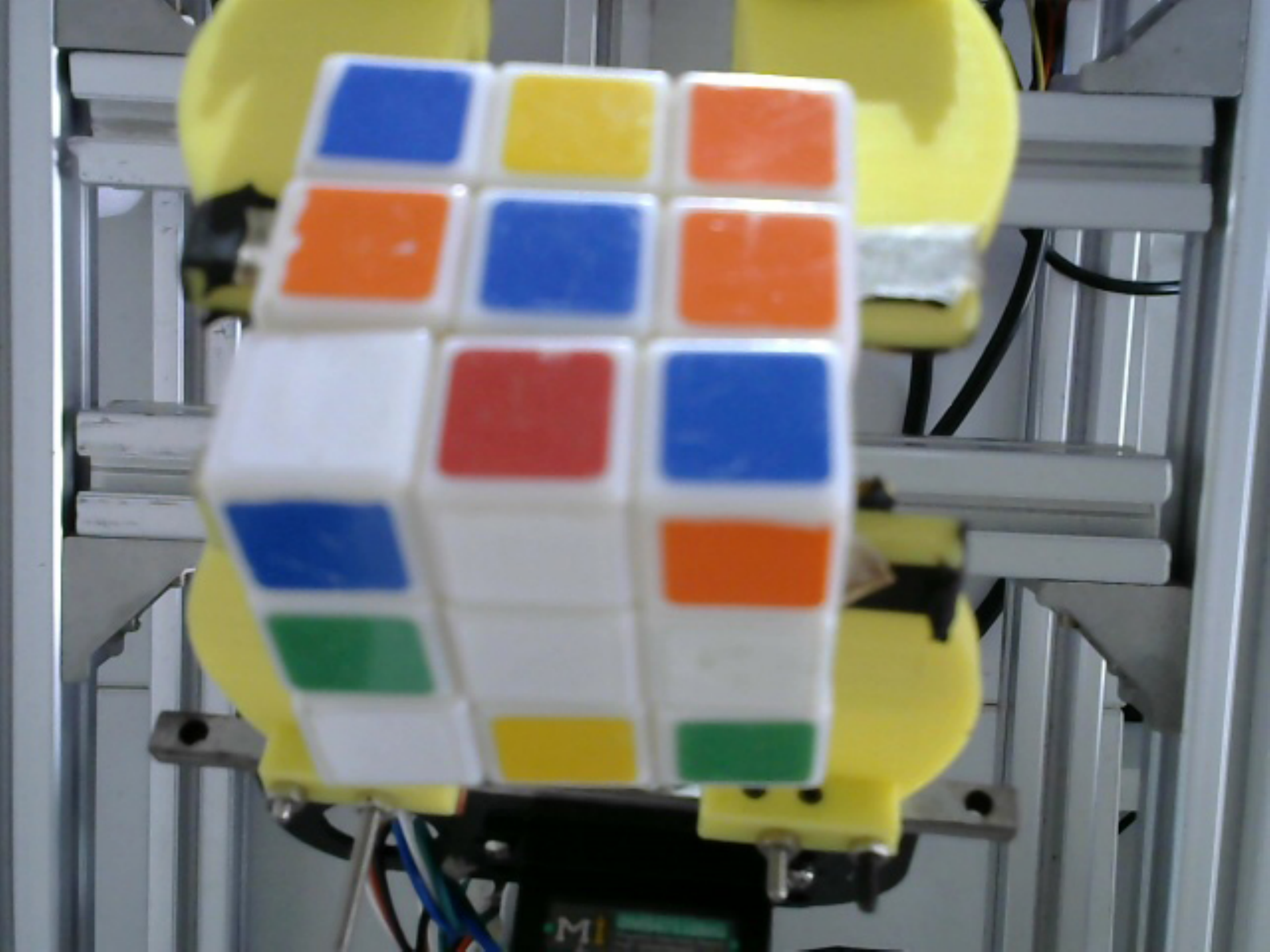}
    \includegraphics[height=1.5cm]{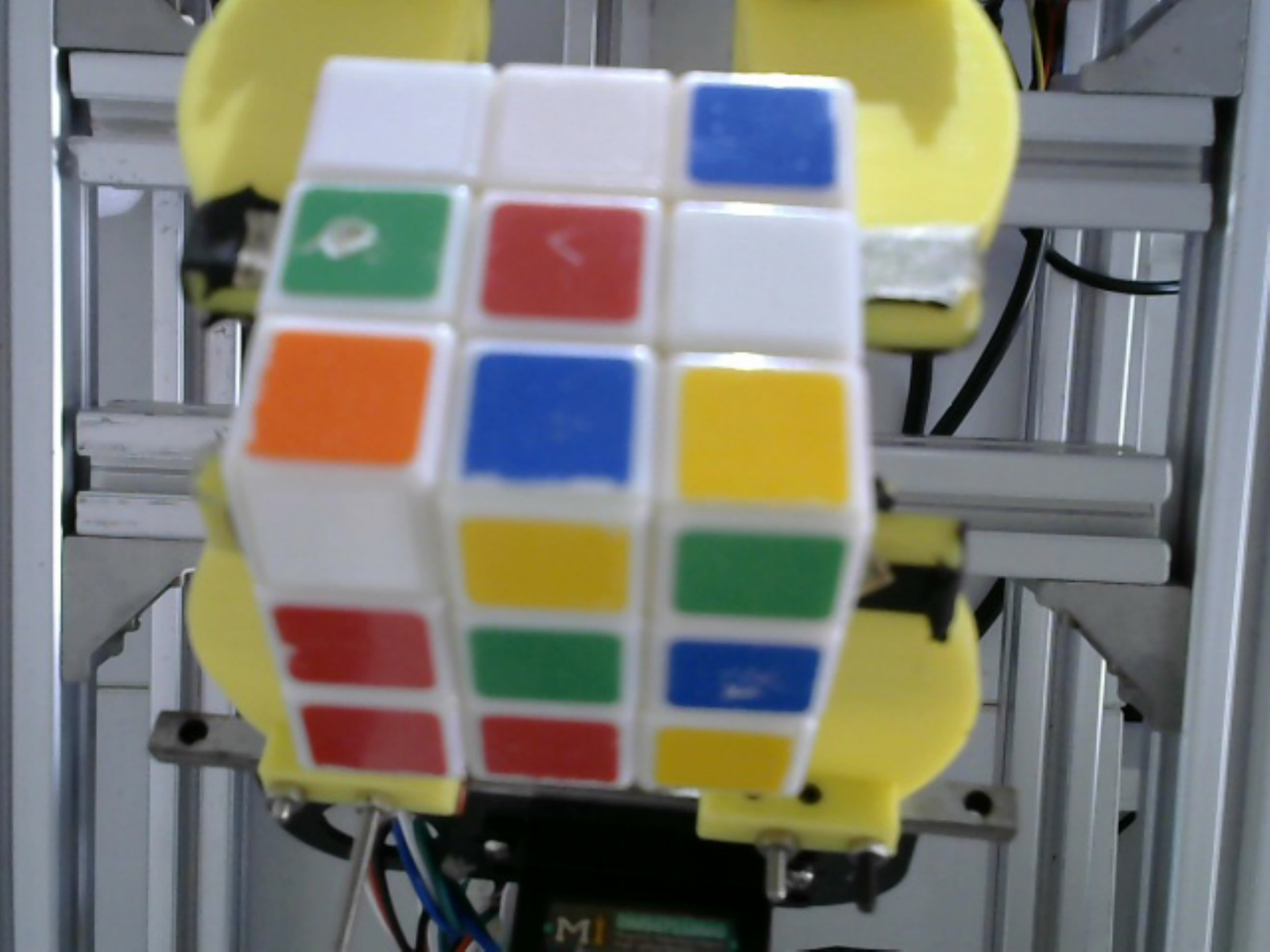}
}

\subfigure[cube B illuminated from above]{
    \centering
    \includegraphics[height=1.5cm]{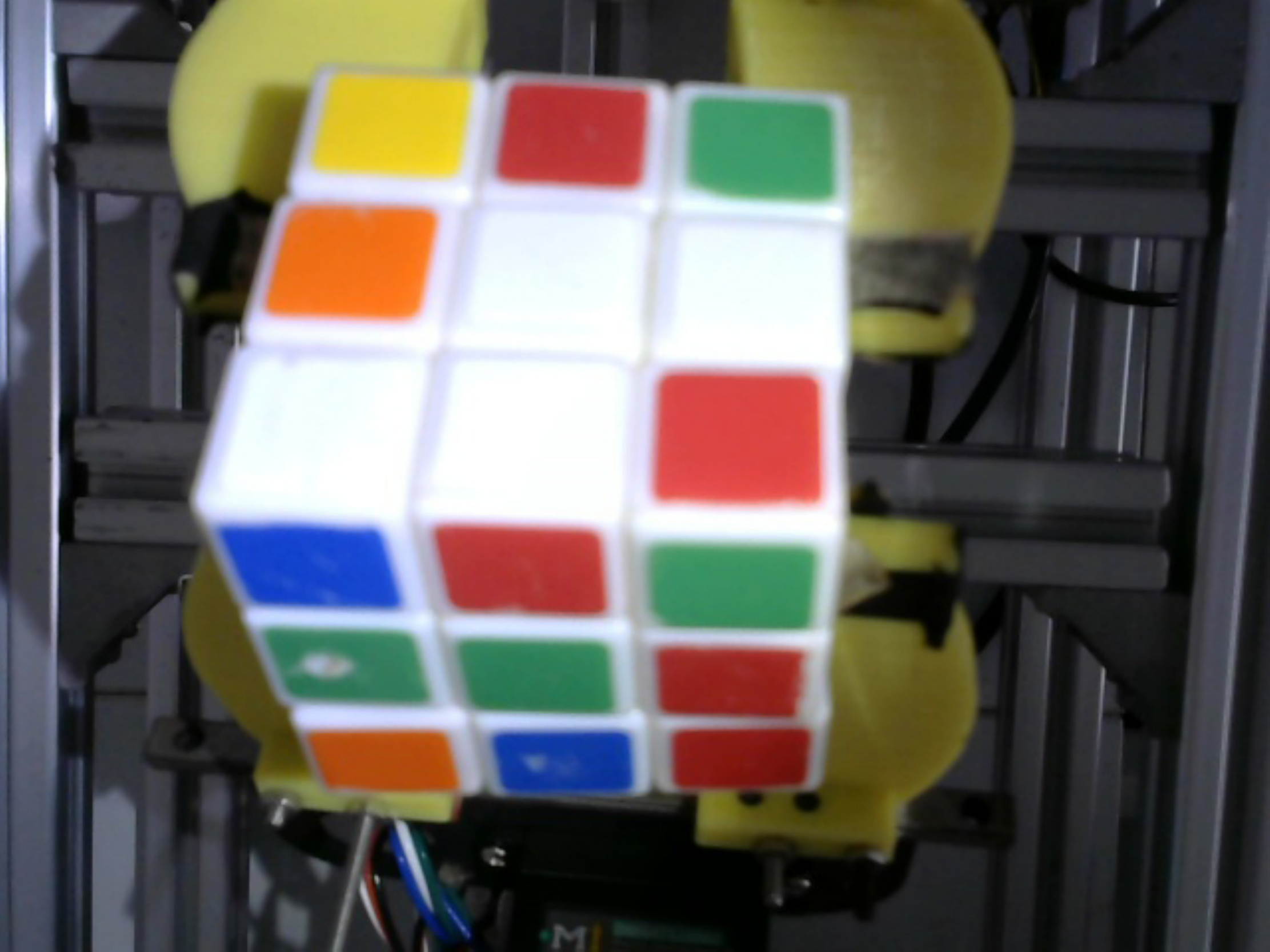}
    \includegraphics[height=1.5cm]{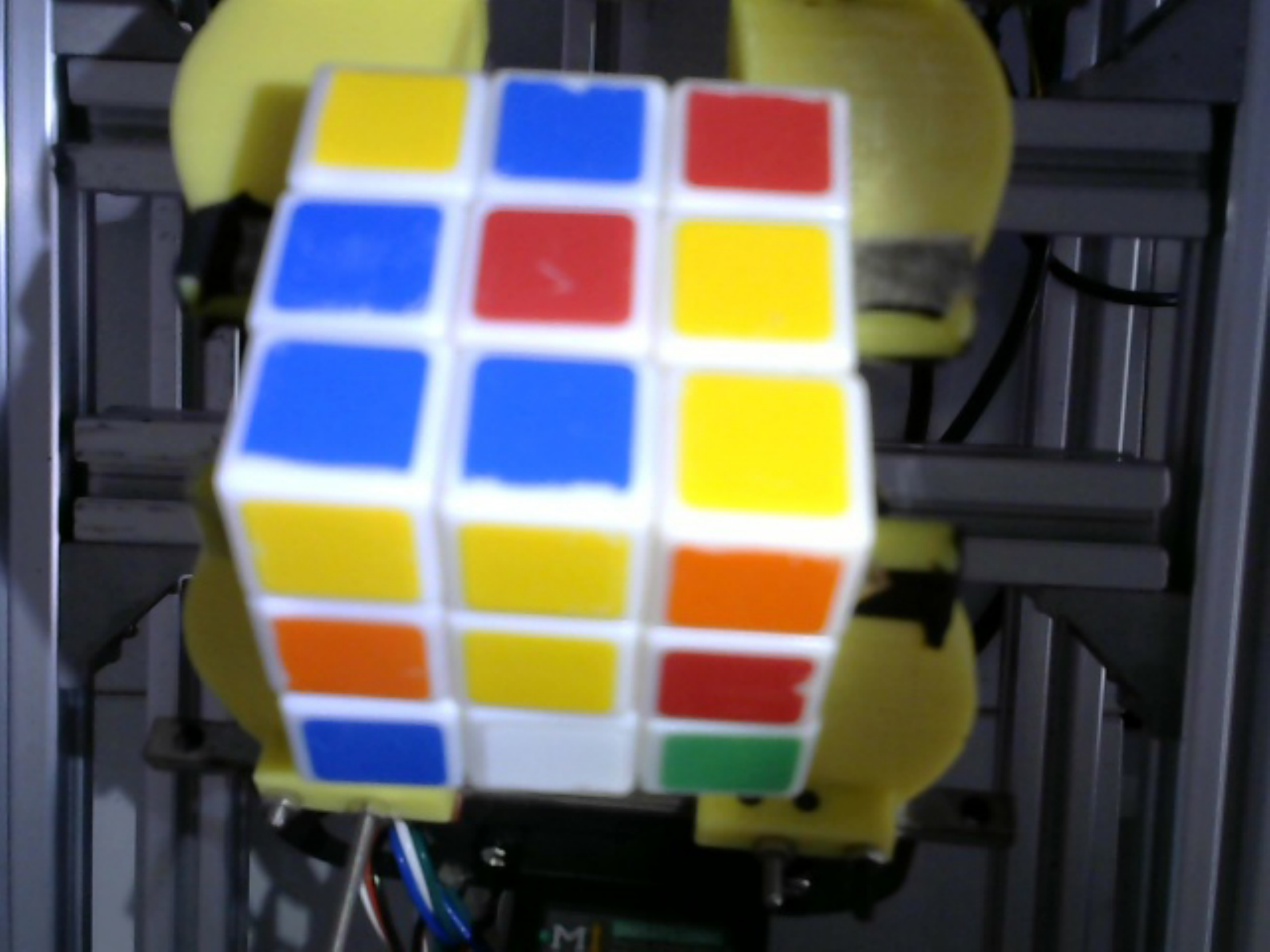}
    \includegraphics[height=1.5cm]{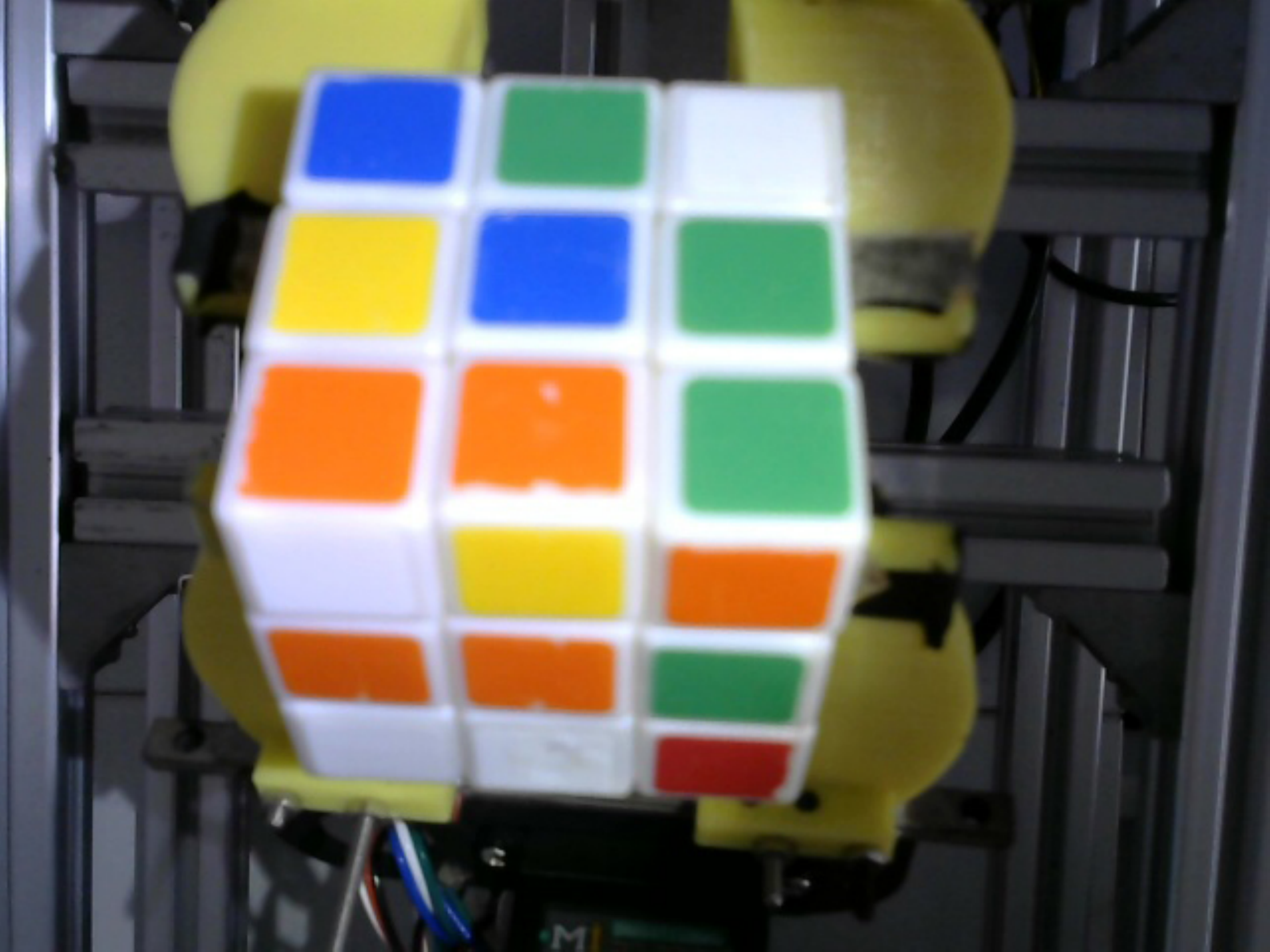}
}

\caption{Examples in Rubik's Cube Dataset}
\label{fig_examples}
\end{figure}

The entire dataset consists of 348 images with the same resolution (640 $\times$ 480),   
  a total of 18792 color block information. The dataset contains the following 5 circumstances: 
  cube A in bright, cube A in dark, cube B in bright, cube B illuminated from right, 
   cube B illuminated from above. A and B are cubes with two different backgrounds. 
   Fig. \ref{fig_examples} shows examples of different circumstances.
\begin{figure}[htbp]
    \centering
    \includegraphics[height=7cm]{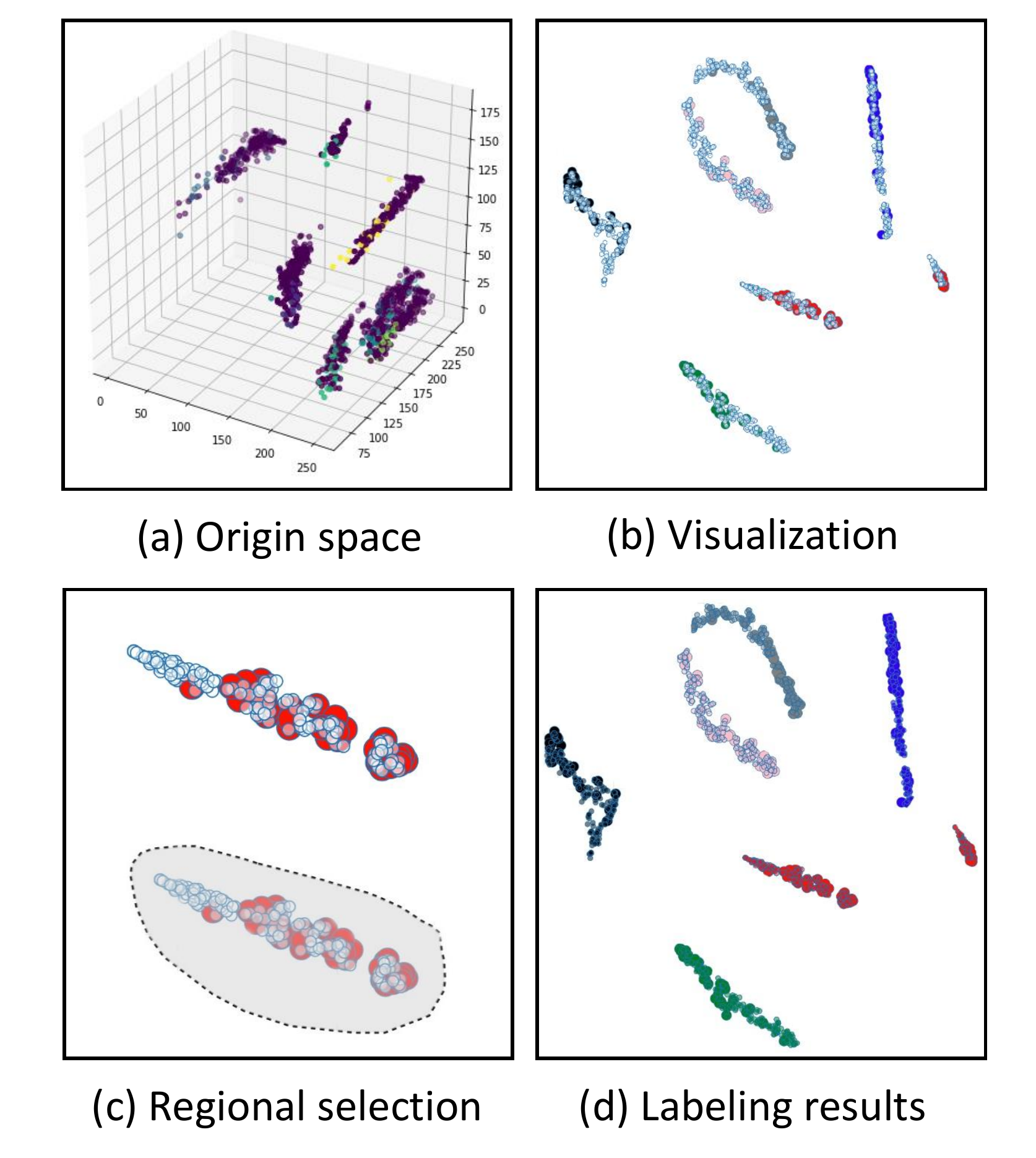}
    \caption{The process of PVIL}
    \label{PVIL}
\end{figure}

\subsection{Label samples}
To get labels of the Rubik’s cube dataset, we adopt a visual labeling method called PVIL labeling \cite{liu2018perceptual}. 
Compared with original handcraft labeling method, the PVIL method is stable in accuracy and efficient 
in time cost. The process of PVIL includes the following 4 steps as shown in Fig. \ref{PVIL}. (a) mapping the origin data into a 
feature subspace. (b) Visualizing feature vector. (c) To judge and select the data which could be 
fallen into one group. (d) The labeling results is obtained.

\section{Experiment}

In this section, we evaluate proposed methods on the Rubik’s Cube dataset introduced above. 
Methods are divided into an offline method and online methods. SB-ELM is an offline method, and 16DHSV features
 are used in SB-ELM. WLHP and DWLP are online methods based on 3DHSV features. In our experiment,
  we use A to E to represent the five circumstances in the Rubik’s cube dataset. 
  Circumstances are arranged according to the order from A to E as follows: cube A in bright, 
  cube A in dark, cube B in bright, cube B illuminated from right, cube B illuminated from above.

\subsection{Accuracy of offline method}
\begin{table}[htbp]
    \begin{center}
    \centering
    \caption{Accuracy of SB-ELM with 16DHSV} \label{tab:elm}
    \begin{tabular}{ccccccc}
      \hline
      samples&A&B&C&D&E&total \\
      \hline
    50   & 85.75         & 96.06         & 82.03        & 78.07         & 76.15         & 84.54         \\
    100  & 97.76         & 97.88         &\textbf{82.39}& 79.17         & 76.86         & 84.47         \\
    150  & 96.94         & 98.88         & 78.52        & 78.64         & 79.04         & 87.02         \\ 
    200  & 96.96         & 98.98         & 82.34        & 80.34         & 78.42         & 87.22         \\
    250  & \textbf{98.80}& \textbf{99.13}& 82.34        & \textbf{81.37}& \textbf{79.64}& 85.03         \\
    300  & 97.54         & 94.56         & 81.33        & 80.10         & 79.18         & \textbf{87.23}\\
      \hline
    \end{tabular}
    \end{center}
\end{table}
We test the accuracy of SB-ELM in single circumstance and mutiple circumstances respectively.
The accuracy of SB-ELM under a single environmental circumstance is high,
 while the accuracy under mutiple circumstances is relatively low. 
 The reasons of the above situation can fall into two aspects:
 (1) It is hard to generalize all the circumstances caused by complex environment. 
 (2) When the brightness of the light changes, most of features of orange and red in 
 the color space will shift, which result in the overlap distribution of orange and red colors.
As we can see in Tab. \ref{tab:elm}, the accuracy of SB-ELM in circumstance A and B trained 
by 250 samples can reach around 99\%. At the same time, the accuracy in mutiple circumstances is between 84\% and 88\%.

\subsection{Accuracy of online methods}
We test the accuracy of online methods in single circumstance and allover circumstances respectively.
 The accuracy of KNN is used as a baseline. Compared with KNN, 
 WLHP has a higher recognition rate because it considers that the center color block feature
  is not distributed in the date center. To explain more specific, if KNN is used to 
  select 8 neighbors of the center color block at a time, some of the 8 neighbors may be misidentified.
   While WLHP reduces the number of neighbors selected, effectively avoiding
   the occurrence of misjudgment. At the same time, the neighbors of neighbors are used as
    the remaining color blocks to ensure the correct number of identifications. It can be seen from
     Tab. \ref{tab:online} that WLHP can get better results in all circumstances than KNN.
\begin{table}[htbp]
\begin{center}
\centering
\caption{Accuracy of online methods with 3DHSV} \label{tab:online}
\begin{tabular}{ccccccc}
    \hline
    Methods &A&B&C&D&E&total \\
    \hline
    KNN  &95.26           &95.11          &85.19          &79.01          &77.68          &86.45 \\
    WLHP &96.22           &95.85          &87.27          &77.70          &80.70          &87.55 \\ 
    WLHP*&97.63           &\textbf{99.85} &\textbf{99.61} &89.86          &94.15          &96.22 \\
    DWLP   &\textbf{100.0}  &99.26          &96.84          &\textbf{93.88} &\textbf{98.15} &\textbf{97.63} \\
    \hline
\end{tabular}
\end{center}
\end{table}

The optimized WLHP is another version of online method, which is denoted as WLHP*. The 
process is as follows: Find the white color block in center color blocks based on the saturation dimension
 of 3DHSV, then identify other white color blocks by the white center color block.
  When identifying other color blocks, the weight of the hue dimension is increased when calculating
 the distance between two features. The shift of color distribution are more sufficiently considered by increasing the
 weight, base on which the discrimination between colors is more obvious.
 The experiment results in Tab. \ref{tab:online} show that the accuracy of WLHP* is 8.67\% better
  than WLHP on average.

Inspired by WLHP*, DWLP makes full use of color information. DWLP utilizes the
distribution characteristics of the six colors in HSV color space to increase the degree of color differentiation.
Color-varying weight matrix is used to calculate the distance between two features 
when performing different label propagation. After the above improvements, DWLP
achieves a good recognition effect in the experiment. As we can see in Tab. \ref{tab:online}, 
the accuracy of DWLP (97.63\%) is the highest accross the entire dataset.

The accuracy of online methods higher than the offline method because the relative 
distribution of the data is stable regardless of light changes, for example,
 the brightness of the orange is always higher than the red one. In CRRC problem, 
 the experiment results has illustated that the online methods, especially DWLP, can perform better in varied environment. As shown in Tab. \ref{tab:elm} and Tab. \ref{tab:online}, 
  the accuracy of SB-ELM accross entire dataset is only around 87\%, while the accuracy of DWLP is around 97\%.

\section{Conclusion}
CRRC is a very important issue in Rubik’s cube robot which also can be treated as a sequential 
manipulation problem. Therefore, CRRC should be deep analyzed. In this paper, we point out the color drifting
 problem of CRRC. Furthermore, we construct a dataset to compare our proposed offline 
 and online methods, which illustrates the effectiveness of online methods. Meanwhile, 
 to verify our viewpoint, we design a Rubik’s cube robot which works in real-world environment. 
 The experimental results are satisfactory by most users who unknowns the operational
  principle of our Rubik’s cube robot.

\bibliographystyle{IEEEbib}
\bibliography{mypaper}

\end{document}